# OUTLIER DETECTION ALGORITHM FOR CIRCLE FITTING

Ahmet Gökhan POYRAZ

Doğu Pres R&D Center, 1610, Bursa, Turkey

Author's e-mail: gokhanpoyraz@dogupres.com, agpoyraz@gmail.com

**Abstract**

Circle fitting methods are extensively utilized in various industries, particularly in quality control processes and design applications. The effectiveness of these algorithms can be significantly compromised when the point sets to be predicted are noisy. To mitigate this issue, outlier detection and removal algorithms are often applied before the circle fitting procedure. This study introduces the Polar Coordinate-Based Outlier Detection (PCOD) algorithm, which can be effectively employed in circle fitting applications. In the proposed approach, the point set is first transformed into polar coordinates, followed by the calculation of both local and global standard deviations. Outliers are then identified by comparing local mean values with the global standard deviation. The practicality and efficiency of the proposed method are demonstrated by focusing on the high-precision diameter measurement of industrial washer parts. Images from a machine vision system are processed through preprocessing steps, including sub-pixel edge detection. The resulting sub-pixel edge points are then cleaned using the proposed outlier detection and removal algorithm, after which circle fitting is performed. A comparison is made using ten different circle fitting algorithms and five distinct outlier detection methods. The results indicate that the proposed method outperforms the other approaches, delivering the best performance in terms of accuracy within the dataset, thereby demonstrating its potential for enhancing circle fitting applications in industrial environments.

## 1. INTRODUCTION

With the advancement of modern technology, industrial image processing applications have become increasingly accessible and widely utilized across various sectors. From medicine and healthcare to forestry and food industries, image processing techniques are now integral to numerous domains. One of the most prominent applications of these techniques is in the visual inspection and dimensional analysis of manufactured components. In particular, circle fitting has emerged as a fundamental operation for assessing the geometric integrity of circular objects. In typical inspection pipelines, edge detection is first performed to extract boundary information, followed by a circle fitting algorithm to estimate the diameter of the object with high precision. The estimated diameter is then used to determine whether the product meets the required specifications. Although circle fitting is a well-established task in the literature, the increasing demand for high-precision manufacturing has highlighted the importance of more robust and accurate fitting algorithms. Numerous methods have been proposed in the literature to address the circle fitting problem. Among these, the Least Squares Fitting (LSF) approach [1] is the most widely adopted due to its simplicity and efficiency. Several studies [2, 3, 4] have successfully employed LSF to solve various measurement tasks. In addition to LSF, alternative algorithms such as Pratt [5], Taubin [6], RANSAC [7], IRLS [8], HyperLS [9], M-Estimator [10], LMedS [11], Total Least Squares (TLS) [12], and EDCircle [13] have been proposed and applied in different contexts. While some of these methods are designed to be more robust against noise and outliers, others prioritize accuracy in low-noise environments.

The Least Squares Fitting (LSF) method is widely favored in industrial applications due to its simplicity and ease of implementation. However, depending on the specific characteristics of the measurement task, different priorities such as computational speed or precision may become more critical, thereby necessitating the use of alternative algorithms. In some scenarios, hardware-based enhancements are also employed to improve image quality, which in turn facilitates more accurate and stable measurement conditions. Nevertheless, industrial environments often introduce unavoidable noise or undesired artifacts on the surface of the components, which can adversely affect the performance of the fitting process. Particularly during the edge detection stage, spurious or irrelevant points may be included in the dataset, leading to distorted results in the subsequent circle fitting phase. To address this issue, two common strategies are adopted: either employing robust iterative fitting algorithms or applying outlier removal techniques to eliminate inconsistent edge points before fitting. Among the most frequently used approaches for outlier removal algorithm are Median Absolute Deviation (MAD) [14], Z-Score [15], DBSCAN [16], Local Outlier Factor (LOF) [17], and Percentile-based filtering [18], each offering different levels of robustness depending on the noise characteristics and spatial distribution of the data.

In this study, a comprehensive comparison of circle fitting algorithms has been conducted to evaluate their performance in the context of industrial image processing. Performance analyses were carried out using real images acquired from a machine vision system developed based on the most commonly used algorithms in the literature. The evaluation includes both standalone implementations of the fitting algorithms and combinations with outlier removal techniques to observe their effect on accuracy and robustness. Furthermore, a new outlier detection algorithm based on polar coordinates has been introduced, which enhances the overall performance of circle fitting by effectively eliminating inconsistent edge points before the fitting process. This method integrates localized evaluation with a global threshold, improving the accuracy of the circle fitting results, particularly in noisy and contaminated industrial environments.

The main contributions of this study are as follows:

- A systematic comparison of existing circle fitting algorithms based on their mathematical properties and empirical performance.
- An investigation into the influence of outlier removal techniques, including the newly proposed polar coordinate-based outlier detection(PCOD) algorithm, on the accuracy and stability of circle fitting results
- A practical performance evaluation framework tailored to the requirements and challenges of industrial image processing applications.

## 2. THEORETICAL BACKGROUND

The circle fitting process in this study begins with a grayscale image $I$ containing a circular or circular-like shape. To achieve precise boundary localization, a subpixel edge detection technique is employed, resulting in a set of edge coordinates $(x_i, y_i)$ where $i = 1, 2, \ldots, n$. These edge points represent the estimated contour of the target shape with high spatial resolution. The primary objective is to infer the parameters of a circle that best describes this set of points. This is formulated as an optimization problem where a fitting model aims to minimize the deviation between the detected points and the ideal circular geometry. Depending on the algorithmic approach, the fitting criterion may rely on algebraic or geometric distances. In the subsequent sections, multiple circle fitting algorithms are presented and compared in terms of their mathematical formulation, accuracy, and robustness.

While certain fitting algorithms are inherently robust against outliers due to their iterative or consensus-based nature, others are highly sensitive to noisy or irregular edge data. Therefore, in addition to evaluating the fitting algorithms themselves, this study also incorporates a comparative analysis of outlier removal techniques. Outlier removal, in this context, refers to the identification and exclusion of anomalous edge points that deviate significantly from the expected circular contour. By eliminating such points prior to the fitting process, the overall accuracy and stability of the circle estimation are substantially improved.

### 2.1 Circle Fitting Algorithms

#### 2.1.1 Least Squares Fitting

Least Squares Circle Fitting (LSF) is one of the most commonly used approaches for estimating the parameters of a circle from a set of edge points. The primary objective is to identify the center coordinates $(a, b)$ and the radius $R$ of a circle that best fits the given $n$ data points $\{(x_i, y_i)\}_{i=1}^{n}$. The ideal circle is defined by the equation:

$$(x - a)^2 + (y - b)^2 = R^2$$

In the geometric form, the LSF method minimizes the sum of squared residuals between the observed distances and the radius:

$$\min_{a,b,R} \sum_{i=1}^{n} \left( \sqrt{(x_i - a)^2 + (y_i - b)^2} - R \right)^2$$

However, due to the nonlinearity introduced by the square root, solving this formulation directly requires iterative numerical optimization, which can be computationally expensive. Therefore, in practical applications, a more tractable algebraic least squares formulation is often preferred. By expanding the circle equation, the implicit form can be written as:

$$x^2 + y^2 + Dx + Ey + F = 0$$

Here, the parameters are related to the original circle parameters as:

$$D = -2a, \quad E = -2b, \quad F = a^2 + b^2 - R^2$$

The fitting objective then becomes minimizing the algebraic error:

$$\min_{D,E,F} \sum_{i=1}^{n} \left( x_i^2 + y_i^2 + Dx_i + Ey_i + F \right)^2$$

This expression can be solved efficiently using linear least squares methods. Defining matrix $A$ and vector $B$ as:

$$A = \begin{bmatrix} x_1 & y_1 & 1 \\ x_2 & y_2 & 1 \\ \vdots & \vdots & \vdots \\ x_n & y_n & 1 \end{bmatrix}, \quad B = \begin{bmatrix} -(x_1^2 + y_1^2) \\ -(x_2^2 + y_2^2) \\ \vdots \\ -(x_n^2 + y_n^2) \end{bmatrix}$$

The optimal solution $[D \quad E \quad F]^T$ is obtained using the normal equation:

$$\begin{bmatrix} D \\ E \\ F \end{bmatrix} = (A^T A)^{-1} A^T B$$

Once the parameters are estimated, the circle's center and radius are recovered as:

$$a = -\frac{D}{2}, \quad b = -\frac{E}{2}, \quad R = \sqrt{a^2 + b^2 - F}$$

### 2.1.2 Pratt Circle Fitting

Pratt's method is an enhanced algebraic circle fitting technique that addresses the limitations of traditional least squares approaches by introducing a geometric constraint. Unlike the standard algebraic method, which minimizes the total algebraic error without enforcing circularity, Pratt's formulation ensures that the fitted conic truly represents a circle, thereby improving numerical robustness and reducing bias. The general implicit equation of a circle is given by:

$$x^2 + y^2 + Dx + Ey + F = 0$$

To fit a circle to the set of $n$ edge points $(x_i, y_i)$, the method defines a parameter vector:

$$\theta = \begin{bmatrix} D \\ E \\ F \end{bmatrix}$$

A corresponding design matrix $A$ is formed using the data points:

$$A = \begin{bmatrix} x_1 & y_1 & 1 \\ x_2 & y_2 & 1 \\ \vdots & \vdots & \vdots \\ x_n & y_n & 1 \end{bmatrix}$$

The circle fitting problem is then formulated as a constrained minimization:

$$\min_\theta |A\theta + b|^2 \quad \text{subject to} \quad \theta^T C \theta = 1$$

Here, $C$ is the constraint matrix, defined to enforce the circularity condition. This leads to a generalized eigenvalue problem:

$$A^T A \theta = \lambda C$$

The solution $\theta$ is the eigenvector corresponding to the smallest positive eigenvalue $\lambda$. Once $D$, $E$, and $F$ are estimated, the geometric parameters of the circle are calculated as:

### 2.1.3 Taubin Circle Fitting

Taubin Circle Fitting is an algebraic fitting method that improves upon the standard least squares and Pratt methods by minimizing a first-order geometric approximation of the squared distance from the data points to the estimated circle. The method reduces geometric bias and avoids the scaling ambiguity of purely algebraic approaches. Similar to other methods, the circle is represented by the general quadratic form:

$$x^2 + y^2 + Dx + Ey + F = 0$$

The key idea in Taubin's approach is to minimize the algebraic distance normalized by the gradient magnitude, which approximates the geometric distance more accurately than the raw algebraic error. The objective function becomes:

$$\min_\theta \frac{|A\theta|^2}{|J\theta|^2}$$

Where:

- $A\theta$ is the vector of algebraic residuals,
- $J\theta$ is the Jacobian-based gradient term,
- $\theta = [D \quad E \quad F]^T$

This results in a generalized eigenvalue problem similar to Pratt's method, but with a different constraint matrix $C_T$:

$$A^T A \theta = \lambda C_T \theta$$

After solving for $\theta$, the original circle parameters are recovered using:

$$a = -\frac{D}{2}, \quad b = -\frac{E}{2}, \quad R = \sqrt{a^2 + b^2 - F}$$

### 2.1.4 RANSAC Circle Fitting

The RANSAC (Random Sample Consensus) algorithm is a widely used robust estimation method designed to deal with datasets contaminated by a high percentage of outliers. Unlike deterministic least squares-based techniques, RANSAC adopts a probabilistic approach that iteratively fits models to randomly selected minimal subsets of data and evaluates their consistency with the overall dataset. In the context of circle fitting, the objective is to estimate the parameters $(a, b, R)$ of the circle defined by the standard implicit equation:

$$(x - a)^2 + (y - b)^2 = R^2$$

The RANSAC algorithm proceeds as follows:

1. Minimal Sampling: At each iteration, a minimal subset of three non-collinear edge points $(x_i, y_i)$ is randomly selected. These points uniquely define a candidate circle.

2. Model Estimation: The circle parameters $(a, b, R)$ are computed analytically from the selected triplet.

3. Inlier Evaluation: The algorithm then assesses the quality of the candidate model by computing the distance of each data point to the estimated circle. The absolute radial distance for each point is given by:

4. Consensus Set Construction: All points for which $d_i < \epsilon$ is a predefined inlier threshold, are considered inliers. The size of this consensus set determines the model's support.

5. Iteration and Selection: This process is repeated for a fixed number of iterations, or until a model with a sufficiently large consensus set is found. The best-fitting model is the one with the highest inlier count.

### 2.1.5 IRLS Circle Fitting

Iteratively Reweighted Least Squares (IRLS) is a robust variant of the classical least squares method, designed to reduce the impact of outliers by assigning adaptive weights to each data point during the fitting process. In circle fitting, the goal is to estimate the center $(a, b)$ and the radius $R$ of a circle that best fits a given set of edge points $(x_i, y_i)$. Unlike traditional least squares, IRLS iteratively updates the contribution of each point based on its residual error. The weighted cost function is defined as:

$$\min_{a,b,R} \sum_{i=1}^{n} w_i \left( \sqrt{(x_i - a)^2 + (y_i - b)^2} - R \right)^2$$

The residual of the $i$-th data point at iteration $k$ is calculated as:

$$r_i^{(k)} = \sqrt{(x_i - a^{(k)})^2 + (y_i - b^{(k)})^2} - R^{(k)}$$

The weight for the next iteration is updated using a robust loss-based rule. A commonly used scheme is:

$$w_i^{(k+1)} = \text{if } \left|r_i^{(k)}\right| \leq \delta, \text{ then } 1, \text{ else } \frac{\delta}{\left|r_i^{(k)}\right|}$$

This formulation allows IRLS to suppress the influence of points with large deviations while preserving the contribution of inliers. The method starts with an initial parameter estimate typically obtained through an algebraic method such as Least Squares Fitting (LSF) and iteratively refines the circle parameters until convergence.

### 2.1.6 HyperLS Circle Fitting

The Hyperaccurate Least Squares (HyperLS) method is a refined algebraic approach for fitting circles to a set of edge points with enhanced numerical precision. Unlike conventional least squares methods, which may produce biased estimates due to unnormalized algebraic error terms, the HyperLS method introduces a constraint to stabilize the solution and avoid trivial results. The general implicit form of a circle is expressed as:

$$x^2 + y^2 + Dx + Ey + F = 0$$

Given a set of $n$ points $(x_i, y_i)$, the goal is to estimate the parameters $D, E, F$ by minimizing the algebraic error function:

$$\min_{D,E,F} \sum_{i=1}^{n} (x_i^2 + y_i^2 + Dx_i + Ey_i + F)^2$$

To avoid trivial solutions $D = E = F = 0$, a normalization constraint is applied:

$$4AC - B^2 = 1$$

Where the statistical moments $A, B, C$, are computed as:

$$A = \langle x^2 \rangle = \frac{1}{n}\sum_{i=1}^{n} x_i^2, \quad B = \langle xy \rangle = \frac{1}{n}\sum_{i=1}^{n} x_i y_i, \quad C = \langle y^2 \rangle = \frac{1}{n}\sum_{i=1}^{n} y_i^2$$

Once the parameters $D, E, F$ are estimated, the center and radius of the circle are recovered as:

$$a = -\frac{D}{2}, \quad b = -\frac{E}{2}, \quad R = \sqrt{a^2 + b^2 - F}$$

### 2.1.7 M-Estimator Circle Fitting

The M-Estimator Circle Fitting method provides a robust alternative to conventional least squares techniques by mitigating the impact of outliers during parameter estimation. Unlike traditional methods that minimize the sum of squared residuals, M-estimators employ a robust loss function $\rho(r_i)$ to downweight the influence of large deviations, thereby increasing the reliability of the fitted circle in the presence of noise or spurious points. The geometric error between the observed point $(x_i, y_i)$ and the estimated circle $(a, b, R)$ is given by:

$$r_i = \sqrt{(x_i - a)^2 + (y_i - b)^2} - R$$

The M-estimator seeks to minimize the cumulative robust cost across all residuals:

$$\min_{a,b,R} \sum_{i=1}^{n} \rho(r_i)$$

A common choice for $\rho$ is the Huber loss function, which behaves quadratically for small residuals and linearly for larger ones, and is defined as:

$$\rho(r) = \begin{cases} \frac{1}{2}r^2, & \text{if } |r| \leq \delta \\ \delta\left(|r| - \frac{1}{2}\delta\right), & \text{otherwise} \end{cases}$$

The minimization is typically solved via Iteratively Reweighted Least Squares (IRLS). At each iteration $k$, weights are updated according to:

$$w_i^{(k+1)} = \frac{\frac{d}{dr}\rho\left(r_i^{(k)}\right)}{r_i^{(k)}}$$

These weights are then used in a weighted least squares update to refine the parameters $a, b, R$. The procedure is repeated until convergence is achieved.

### 2.1.8 LMedS Circle Fitting

The Least Median of Squares (LMedS) method is a robust statistical approach for fitting geometric shapes, particularly effective in datasets contaminated with significant outliers. Unlike conventional least squares methods that aim to minimize the sum of squared residuals, the LMedS algorithm minimizes the median of these squared residuals. This distinction allows the method to tolerate up to 50% outlier contamination without significant degradation in performance. Given a set of points $\{(x_i, y_i)\}_{i=1}^{n}$, the goal is to estimate the parameters of a circle defined by its center $(a, b)$ and radius $R$. For each data point, the residual between the observed location and the candidate circle is computed as:

$$r_i = \sqrt{(x_i - a)^2 + (y_i - b)^2} - R$$

Rather than minimizing the total error across all points, LMedS seeks the parameter set that minimizes the median of these squared residuals. The objective function is therefore expressed as:

$$\min_{a,b,R} \text{median}\,(r_1^2, r_2^2, \ldots, r_n^2)$$

This formulation is particularly advantageous in real-world applications where noise or incorrect edge detections may introduce large deviations in a subset of the data. By focusing on the median rather than the mean, LMedS reduces sensitivity to such anomalies. In practice, the LMedS algorithm operates by randomly sampling minimal subsets from the data typically three non-collinear points which are sufficient to define a unique circle. For each sampled triplet, a candidate circle is fitted, and the squared residuals of all data points with respect to this circle are computed. The model that yields the smallest median of these residuals is selected as the final solution.

### 2.1.9 TLS Circle Fitting

The Total Least Squares (TLS) method offers a refined solution for fitting a circle to a set of edge points, particularly when measurement errors are present in both coordinate directions. Unlike the conventional least squares approach, which assumes that all deviations are confined to the dependent variable, TLS considers the possibility that errors may exist in both $x$ and $y$, providing a more symmetric and robust estimation framework. Let $\{(x_i, y_i)\}_{i=1}^{n}$ denote the set of observed edge points extracted from a subpixel detection algorithm. The implicit equation of a circle can be expressed as:

$$x^2 + y^2 + Dx + Ey + F = 0$$

Each point approximately satisfies this equation:

$$x_i^2 + y_i^2 + Dx_i + Ey_i + F \approx 0$$

This can be rewritten in matrix form as a homogeneous linear system:

$$A \cdot \theta \approx 0$$

Here, the design matrix $A$ and parameter vector $\theta$ are defined as:

$$A = \begin{bmatrix} x_1 & y_1 & 1 & -(x_1^2 + y_1^2) \\ x_2 & y_2 & 1 & -(x_2^2 + y_2^2) \\ \vdots & \vdots & \vdots & \vdots \\ x_n & y_n & 1 & -(x_n^2 + y_n^2) \end{bmatrix}, \quad \theta = \begin{bmatrix} D \\ E \\ F \\ 1 \end{bmatrix}$$

The optimal parameter vector $\theta$ corresponds to the right singular vector associated with the smallest singular value of $A$, which is obtained via Singular Value Decomposition (SVD). Once the coefficients $D, E,$ and $F$ are estimated, the circle's center $(a, b)$ and radius $R$ are recovered using:

$$a = -\frac{D}{2}, \quad b = -\frac{E}{2}, \quad R = \sqrt{a^2 + b^2 - F}$$

This approach is particularly suitable when the measurement process introduces noise along both axes, as it provides a geometrically consistent solution by minimizing the orthogonal distance from the data points to the estimated circle.

### 2.1.10 EDCircle Circle Fitting

The Eigenvalue Decomposition Circle Fitting (EDCircle) method is a numerically stable algebraic technique used for estimating the parameters of a circle that best fits a set of 2D points. Unlike iterative geometric methods, EDCircle leverages the properties of eigenvalue decomposition to derive the circle parameters in a closed-form and computationally efficient manner. This method avoids non-linear optimization and is particularly suitable for large-scale or real-time applications. Let the general equation of a circle be expressed in algebraic form:

$$x^2 + y^2 + Dx + Ey + F = 0$$

Given a set of nnn points $\{(x_i, y_i)\}_{i=1}^{n}$, we define the design matrix $A$ as:

$$A = \begin{bmatrix} x_1 & y_1 & 1 & -(x_1^2 + y_1^2) \\ x_2 & y_2 & 1 & -(x_2^2 + y_2^2) \\ \vdots & \vdots & \vdots & \vdots \\ x_n & y_n & 1 & -(x_n^2 + y_n^2) \end{bmatrix}$$

Let the unknown parameter vector be:

$$\theta = \begin{bmatrix} D \\ E \\ F \\ 1 \end{bmatrix}$$

The objective is to find the best fit by solving the linear system:

$$A\theta = 0$$

This homogeneous linear system is typically solved using Singular Value Decomposition (SVD) or eigenvalue decomposition of the scatter matrix $A^T A$. The solution vector $\theta$ corresponds to the eigenvector associated with the smallest eigenvalue. Once $D$, $E$, and $F$ are obtained, the center $(a, b)$ and radius $R$ of the circle are computed as:

$$a = -\frac{D}{2}, \quad b = -\frac{E}{2}, \quad R = \sqrt{a^2 + b^2 - F}$$

### 2.2 Outlier Removal Algorithms
#### 2.2.1 Median Absolute Deviation Based Outlier Removal

In the context of robust circle fitting, the Median Absolute Deviation (MAD) is used to eliminate edge points that deviate significantly from the estimated circular contour. Given a set of $n$ detected edge points $\{(x_i, y_i)\}_{i=1}^n$, and a preliminary fitted circle, the Euclidean distance between each point and the estimated circle boundary is computed as the residual $r_i$. Let these residuals be denoted by $\{r_i\}_{i=1}^n$. The MAD is then defined as:

$$\text{MAD} = \text{median}(|r_i - \text{median}(r)|)$$

To detect outliers, a normalized score $z_i$ is calculated for each residual:

$$z_i = \frac{|r_i - \text{median}(r)|}{\text{MAD} + \varepsilon}$$

An edge point $(x_i, y_i)$ is considered an outlier if the following condition is satisfied:

$$|z_i| > \tau$$

where $\tau$ is a user defined threshold, typically between 2.5 and 3.5. This MAD based method is particularly robust under non-Gaussian noise or partial occlusion, and effectively suppresses the influence of outliers prior to final model estimation.

#### 2.2.2 Z-Score Based Outlier Removal

The Z-score method is a classical statistical approach for detecting outliers by measuring how far each data point deviates from the mean in terms of standard deviations. In the context of circle fitting, this method can be applied to the residuals computed between the observed edge points and the preliminary estimated circle. Given a set of edge points $\{(x_i, y_i)\}_{i=1}^n$, we first compute the Euclidean residuals $\{r_i\}_{i=1}^n$ as the distances from each point to the initial fitted circle. The Z-score $z_i$ for each residual is defined as:

$$z_i = \frac{r_i - \mu_r}{\sigma_r}$$

where:

- $\mu_r = \frac{1}{n}\sum_{i=1}^n r_i$ is the sample mean of the residuals,
- $\sigma_r = \sqrt{\frac{1}{n}\sum_{i=1}^n (r_i - \mu_r)^2}$ is the standard deviation of the residuals.

A residual is flagged as an outlier if:

$$|z_i| > \tau$$

#### 2.2.3 Density-Based Outlier Removal (DBSCAN)

DBSCAN (Density-Based Spatial Clustering of Applications with Noise) is a density-based clustering algorithm that is also widely used for outlier detection. Unlike statistical methods that rely on distributional assumptions, DBSCAN identifies clusters based on the spatial density of data points and treats low-density regions as outliers. In the context of circle fitting, DBSCAN is applied directly to the edge point coordinates $\{(x_i, y_i)\}_{i=1}^n$ to eliminate isolated or noisy points before fitting. DBSCAN requires two user-defined parameters:

- $\varepsilon$: the maximum distance between two points to be considered neighbors,

- $minPts$: the minimum number of neighbors required to form a dense region.

A point $(x_i, y_i)$ is classified as:

- a core point if it has at least $minPts$ points within distance $\varepsilon$,
- a border point if it is within distance $\varepsilon$ of a core point but has fewer than $minPts$ neighbors,
- a noise (outlier) point if it is neither a core nor a border point.

Mathematically, the $\varepsilon$-neighborhood of a point is defined as:

$$\mathcal{N}_\varepsilon(x_i, y_i) = \left\{ (x_j, y_j) \;\middle|\; \sqrt{(x_j - x_i)^2 + (y_j - y_i)^2} \leq \varepsilon \right\}$$

The condition for a point to be a core point is:

$$|\mathcal{N}_\varepsilon(x_i, y_i)| \geq \text{minPts}$$

Points not satisfying this condition and not reachable from any core point are considered outliers and removed prior to model fitting. Due to its non-parametric nature and robustness to shape variations, DBSCAN is particularly effective in handling non-linear, irregularly distributed noise in edge data.

### 2.2.4 Local Outlier Factor (LOF) Based Outlier Removal

The Local Outlier Factor (LOF) algorithm is a density-based method used to identify local anomalies in a dataset by comparing the local density of a point to that of its neighbors. Unlike global thresholding approaches, LOF considers the varying density of different regions and is thus more effective in detecting outliers in datasets with heterogeneous distributions. Given a set of edge points $\{(x_i, y_i)\}_{i=1}^{n}$, LOF is computed through the following steps:

- *k-distance of a point*

For a given point $(x_i, y_i)$, the distance to its $k$-th nearest neighbor is defined as $k - dist(i)$. The set of $k$-nearest neighbors is denoted:

$$N_k(i) = \left\{ (x_j, y_j) \mid (x_j, y_j) \text{ is among the } k \text{ nearest neighbors of } (x_i, y_i) \right\}$$

- *Reachability Distance*

To account for dense regions, the reachability distance is defined as:

$$\text{reach-dist}_k(i, j) = \max\{k\text{-dist}(j),\, d(i, j)\}$$

where $d(i, j)$ is the Euclidean distance between points $i$ and $j$.

- *Local Reachability Density (LRD)*

The local reachability density of point $i$ is the inverse of the average reachability distance from its neighbors:

$$\text{lrd}_k(i) = \left( \frac{1}{|N_k(i)|} \sum_{j \in N_k(i)} \text{reach-dist}_k(i, j) \right)^{-1}$$

- *Local Outlier Factor (LOF)*

Finally, the LOF score for point $i$ is calculated as the average ratio of the LRDs of its neighbors to its own LRD:

$$\text{LOF}_k(i) = \frac{1}{|N_k(i)|} \sum_{j \in N_k(i)} \frac{\text{lrd}_k(j)}{\text{lrd}_k(i)}$$

A point is considered an outlier if $\text{LOF}_k(i) > \tau$, where $\tau$ is typically in the range 1.5–2.0. LOF is particularly effective for identifying outliers in non-uniformly distributed data, making it well-suited for real-world edge detection scenarios in circle fitting applications.

### 2.2.5 Percentile-Based Outlier Removal

The percentile-based outlier removal method is a non-parametric statistical technique used to eliminate extreme values by identifying data points that lie outside a specified percentile range. Unlike Z-score or MAD, this method does not assume any particular distribution and is particularly effective in handling asymmetric or skewed datasets.

Given a set of residuals $\{r_i\}_{i=1}^n$, which represent the distances of the edge points $\{(x_i, y_i)\}_{i=1}^n$ to a preliminary circle model, the method defines two cut-off thresholds based on lower and upper percentiles corresponding to ±2 standard deviations of a normal distribution. These thresholds, approximately the 2.275th and 97.725th percentiles, retain 95.45% of the data, thus ensuring a statistically grounded balance between eliminating outliers and preserving valid information.

Let:

- $P_\alpha$: the lower percentile (≈ 2.275),
- $P_{1-\alpha}$: the upper percentile (≈ 97.725).

The interval for acceptable residual values is defined as:

$$P_\alpha \leq r_i \leq P_{1-\alpha}$$

Any residual $r_i$ that falls outside this range is considered an outlier and removed from the dataset prior to final model fitting:

$$r_i < P_\alpha \quad \text{or} \quad r_i > P_{1-\alpha}$$

This statistically justified selection ensures consistency with the properties of the normal distribution, while maintaining computational efficiency and robustness against heavy-tailed or anomalous data distributions.

## 3. EXPERIMENTAL SETUP
### 3.1 Image Acquisition System

In industrial applications, camera-based control systems are typically divided into two categories: fixed-station systems and mobile systems. For workpieces with appropriate geometries, such as disc-shaped parts, rotating glass systems are commonly used. In this study, the PSG-1500 Glass Dial Sorting Machine was employed specifically for image acquisition. Although this machine is generally equipped with proprietary closed-loop algorithms that measure the diameters of parts, it was used solely for image capturing in this research to allow for a more stable comparison. The system is capable of providing images from four distinct stations. However, for measurement purposes, only the first station was selected, where the background image is white, and the part image is black. This particular configuration was chosen to ensure clearer, more controlled images. The process begins with workpieces being fed onto the rotating glass platform through a vibration system, causing multiple parts to fall sequentially onto the glass disc. As each part passes beneath the measurement camera, an image is captured. After completing a full rotation on the glass platform, the parts are transferred into a container using a blowing mechanism. This process ensures that the parts are transferred quickly and without mixing, which is crucial for both image acquisition and subsequent measurement. The equipment used for this system includes a 4 MP industrial camera, which captures high-resolution images of the workpieces, ensuring sufficient detail for accurate measurements. Additionally, a telecentric lens is used, providing a constant working distance to maintain measurement accuracy. This lens was manually adjusted to the required position, as telecentric lenses are designed to keep the working distance fixed. Similarly, the backlight illumination was manually adjusted to the optimal position to enhance the contrast and clarity of the images, further improving the quality of the data acquired. By ensuring that the working distance and lighting conditions remain optimal, the system guarantees the production of stable, high-quality images for further analysis. This setup is vital for ensuring the reliability and stability of the measurement system, which is essential for the accurate performance of the outlier detection algorithm in circle fitting tasks.

### 3.2 Dataset

The most common real-world application of the circle fitting problem is in the measurement of disc-shaped parts, which is why a dataset consisting of 45 disc-shaped workpieces has been selected for this study (Figure 1). The discs in this dataset are used as motor injector setting discs, which have significantly higher precision tolerances compared to regular discs. Specifically, they have a tolerance of 23.7 ± 0.1 mm. The high precision of these workpieces is critical in industrial settings where they undergo serial inspection, making the performance of the fitting algorithm crucial for accurate measurements. Prior to image acquisition, the 45 workpieces underwent a cleaning process to remove any contaminants or debris that could interfere with the measurement. After cleaning, the parts were passed through the PSG-1500 Glass Dial Sorting Machine, which was used to capture the images for analysis. To obtain highly accurate real-world measurements of the parts, a Coordinate Measurement Machine (CMM) was utilized. The CMM provides measurements with an approximate accuracy of 1 μm, ensuring that the true dimensions of the workpieces were precisely determined in millimeters. As the main objective of this study is to compare the performance of circle fitting algorithms, only the outer diameter values of the workpieces were used for analysis. Measuring the inner diameter would require additional preprocessing steps and may introduce repetition in the results. Therefore, the inner diameter measurements were not considered in this study to maintain the focus on the outer diameter fitting and avoid unnecessary complexity.

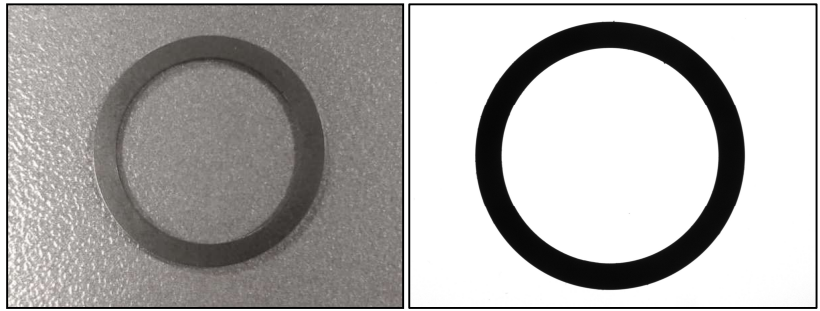

**Figure 1.** Working part and its captured image

## 4. MEASUREMENT METHODOLOGY
### 4.1 Preprocessing

Real-world industrial conditions often differ from controlled laboratory environments and present unexpected challenges. One of the most difficult conditions for camera-based systems in industrial environments is the presence of contamination. Dust, dirt, oil vapors, or any unwanted pollutants from the production environment can accumulate on the camera lens or the workpieces themselves, resulting in significant contamination. This contamination negatively affects both the measurement equipment and the accuracy of the measurement system. Therefore, in order to obtain the best results from industrial environments, it is essential to ensure that the workpieces are thoroughly cleaned, often using non-damaging liquids. In this study, the workpieces were cleaned using special cleaning solutions prior to measurement. The measurement procedures were then carried out following this cleaning step. Despite the cleaning process, even the smallest particles of dust from the environment can stick to the workpieces or the measurement glass, especially in highly sensitive measurement conditions. These particles cause unwanted dark spots on the images, which, in turn, lead to inaccuracies in edge detection. Such inaccuracies may result in the incorrect identification of edges and negatively impact the measurement outcome. To address this issue, a preprocessing step based on connected component labeling was implemented to remove contaminants that were not adhered to the workpieces (Figure 2). The raw image $I$ was processed by performing binarization followed by the complement operation. Connected components were then identified in the image. Assuming that the largest object in the image corresponds to the workpiece, smaller components, likely representing dust or other unwanted noise, were removed. This effectively eliminates the unwanted background contamination. Since the study focuses on outer diameter measurement, only the outer diameter outlier removal and fitting algorithms were applied. In order to avoid the impact of the inner diameter on the results, the cleaned image underwent additional preprocessing steps, including complement, filling, and re-complement operations. These steps produced a fully enclosed image, ensuring that only the outer contour was considered in the subsequent analysis. The subpixel edge detection process was then applied to this image to accurately identify the edges of the workpieces.

### 4.2 Sub-Pixel Edge Detection

Standard edge detection algorithms (such as Canny, Sobel, etc.) typically provide satisfactory results for many applications. However, in precise measurement applications, these conventional edge detection methods often fall short. When measurement accuracy approaches the micron scale, pixel-based algorithms are insufficient to achieve the necessary precision. As a result, researchers have proposed sub-pixel algorithms to better determine the exact point where the edge region intersects a given pixel. Several studies have developed sub-pixel algorithms based on different approaches, including interpolation, moments, fitting, and area-based methods. For instance, in the work of [21], these algorithms are categorized into four classes based on the approach used: interpolation, moments, fitting, and area-based methods [22-30]. One of the most effective sub-pixel edge detection algorithms in the literature, as discussed in [22], was selected for use in this study due to its open-source nature. The chosen method incorporates several sub-parameters, and the default values suggested in the original paper were used for this study: sigma = 1, high threshold = 15, and low threshold = 5. Since the focus of this study is not on evaluating the performance of edge detection methods but rather on assessing the success of the outlier detection algorithm, no modifications were made to the threshold values or alternative sub-pixel edge detection methods were explored. The primary goal was to identify unwanted outliers among the detected edge points, which are crucial for improving the overall measurement accuracy.

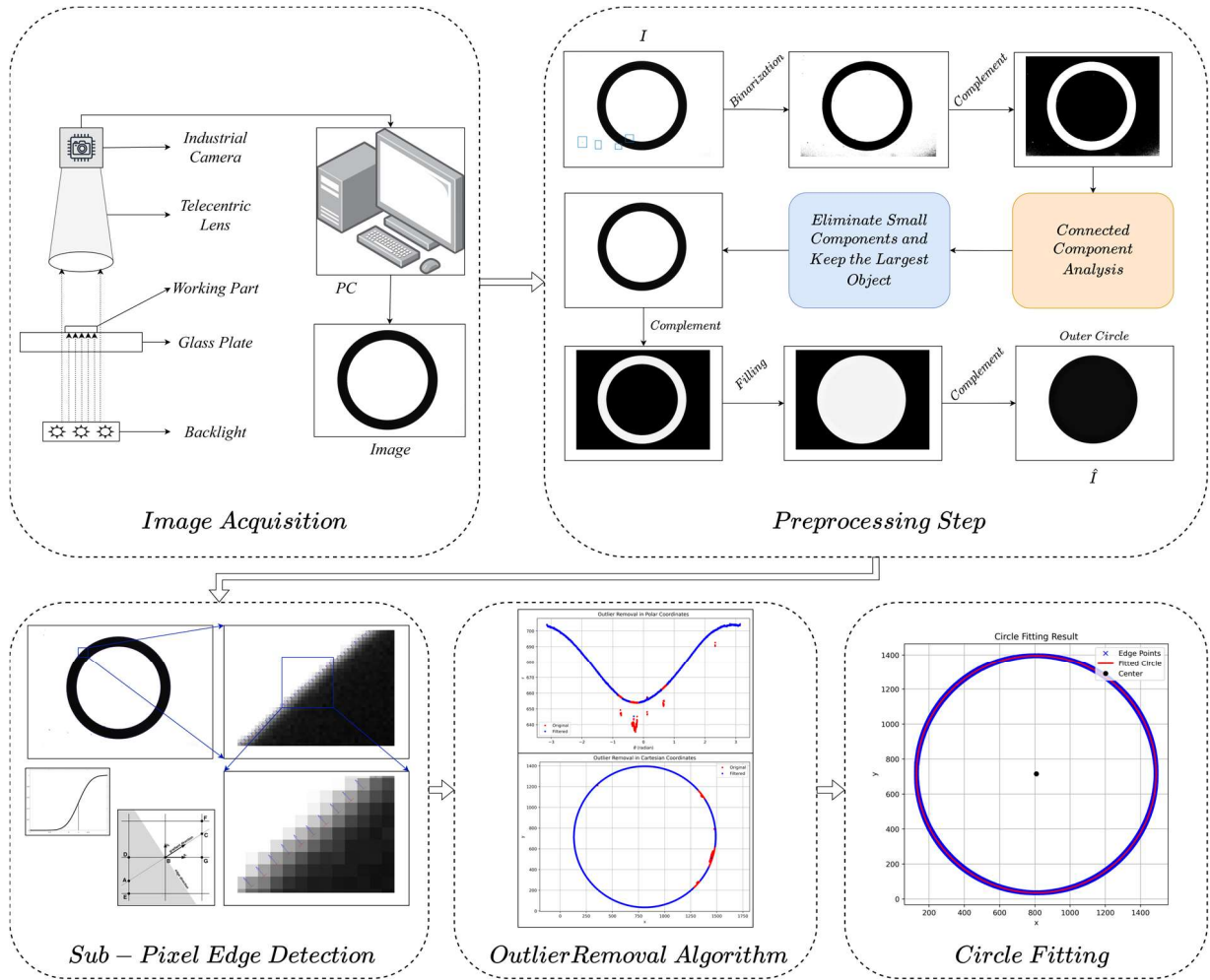

**Figure 2.** Overall workflow of the proposed circular measurement system. The process consists of four main stages: (1) Image Acquisition: An industrial camera with a telecentric lens captures backlit images of the part placed on a glass plate. (2) Preprocessing Step: Binary operations and connected component analysis are performed to isolate the inner and outer circles by eliminating small components and retaining the largest object. (3) Outlier Removal Algorithm: Subpixel edge extraction is applied followed by statistical filtering of outlier edge points. (4) Circle Fitting: Filtered edge points are used to fit an accurate circle model, resulting in a precise estimation of the geometric parameters.

### 4.3 Proposed Outlier Removal Algorithm

In circle fitting applications, the goal is to estimate the best-fitting circle equation based on the detected edge points. However, depending on the application, some unwanted components such as burrs, dust, or dirt may adhere to the workpiece. In such cases, these unwanted regions may also be detected as edge points during the edge detection step. The presence of these unwanted points directly affects the performance of the circle fitting process. While some applications may include these unwanted points in the fitting procedure, in most cases, these points are considered misleading and are therefore removed. To address this issue, outlier removal algorithms are commonly employed. These algorithms are essentially a preprocessing step aimed at cleaning unwanted outliers before the fitting procedure is carried out. Outlier removal helps to improve the accuracy and reliability of the fitting process by eliminating data points that do not correspond to the actual structure of the workpiece. This step is crucial for ensuring that the circle fitting algorithm is not influenced by external factors such as contamination, which could result in erroneous measurements. In summary, outlier removal can be described as a process of cleaning unwanted anomalous points prior to performing the fitting procedure, ensuring that the resulting circle is as accurate as possible, based solely on the relevant edge data.

In this study, an outlier detection strategy is proposed that operates in the polar coordinate space (Figure 3). Unlike conventional statistical techniques, this approach integrates localized window-based evaluation with a globally defined threshold to filter out radius values that do not conform to the expected circular structure. The method proceeds through the following stages:

Given a set of edge points $\{(x_i, y_i)\}_{i=1}^{n}$, the centroid $(\bar{x}_i, \bar{y}_i)$ is calculated as:

$$x_c = \frac{1}{n}\sum_{i=1}^{n} x_i, \quad y_c = \frac{1}{n}\sum_{i=1}^{n} y_i$$

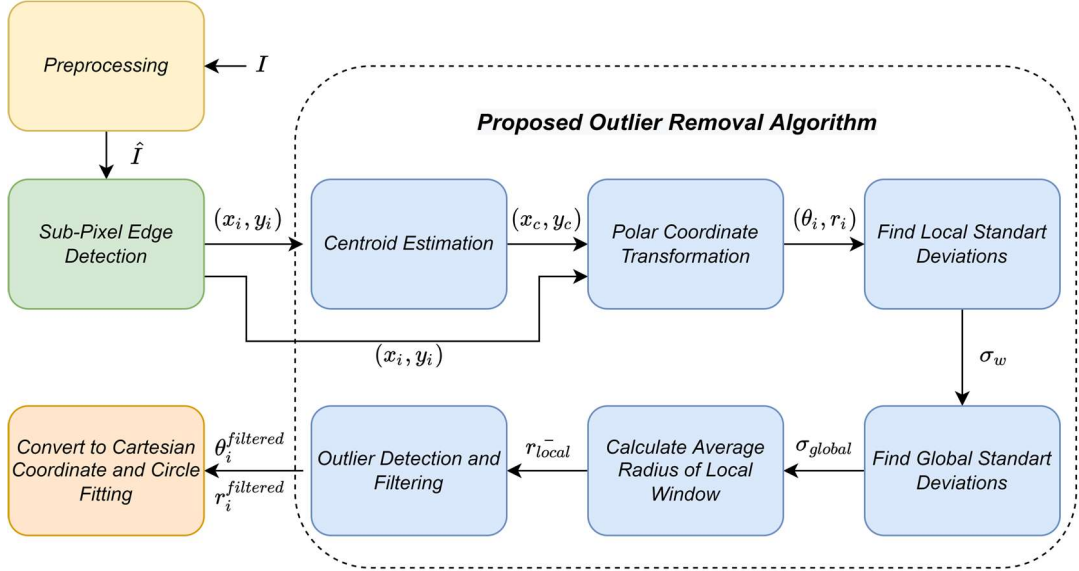

**Figure 3.** Block diagram of the proposed polar coordinate-based outlier removal (PCOD) algorithm.

Using this centroid, all edge points are converted into polar coordinates $(r_i, \theta_i)$ relative to the center:

$$r_i = \sqrt{(x_i - x_c)^2 + (y_i - y_c)^2}, \quad \theta_i = \arctan 2\,(y_i - y_c,\, x_i - x_c)$$

The angular values $\theta_i$ are then sorted in ascending order to maintain continuity in the circular structure. Local standard deviations are computed across radius values within windows of fixed width $w$. These windows are advanced with a constant stride, which controls the step size between consecutive window positions and improves computational efficiency (Figure 4a). The standard deviation for the $j$th window is defined as:

$$\sigma_{w_j} = \sqrt{\frac{1}{w}\sum_{i=j}^{j+w}(r_i - \bar{r}_j)^2}$$

Here, $\bar{r}_j$ is the mean radius within the $j$th window. A global deviation threshold is then computed by taking the median of all window-based standard deviations:

$$\sigma_{global} = \mathrm{median}\left(\{\sigma_{w_j}\}_j\right)$$

For each window segment, the local mean radius is also calculated as:

$$\overline{r_{local}} = \frac{1}{w}\sum_{i=1}^{w} r_i$$

Each individual radius value $r_i$ is evaluated using a global deviation criterion based on the following thresholding rule:

$$|r_i - \overline{r_{local}}| > T \cdot \sigma_{global}$$

If this condition is met, the corresponding point $(x_i, y_i)$ is labeled as an outlier and excluded from further steps (Figure 4b). After filtering, the remaining valid edge points are converted back into Cartesian coordinates using the inverse transformation:

$$x_i^{filtered} = r_i^{filtered} \cdot \cos(\theta_i^{filtered}) + x_c$$

$$y_i^{filtered} = r_i^{filtered} \cdot \sin(\theta_i^{filtered}) + y_c$$

This method offers several advantages in outlier detection for circular structures. By operating in the polar coordinate system, it naturally captures radial inconsistencies and geometric deviations. The combination of local averaging and a global, median-based threshold improves robustness, particularly in datasets where outliers exhibit non-uniform distribution or clustering. The use of the median to compute the global deviation ensures stronger resistance to dense or clustered outliers, which might otherwise distort mean-based estimates. Moreover, the fixed-stride sliding window approach not only enhances computational efficiency but also preserves the angular consistency and geometric integrity of the circular edge patterns. This hybrid strategy ensures that the filtering process remains both statistically reliable and geometrically faithful, contributing to its effectiveness in accurately identifying and removing outliers.

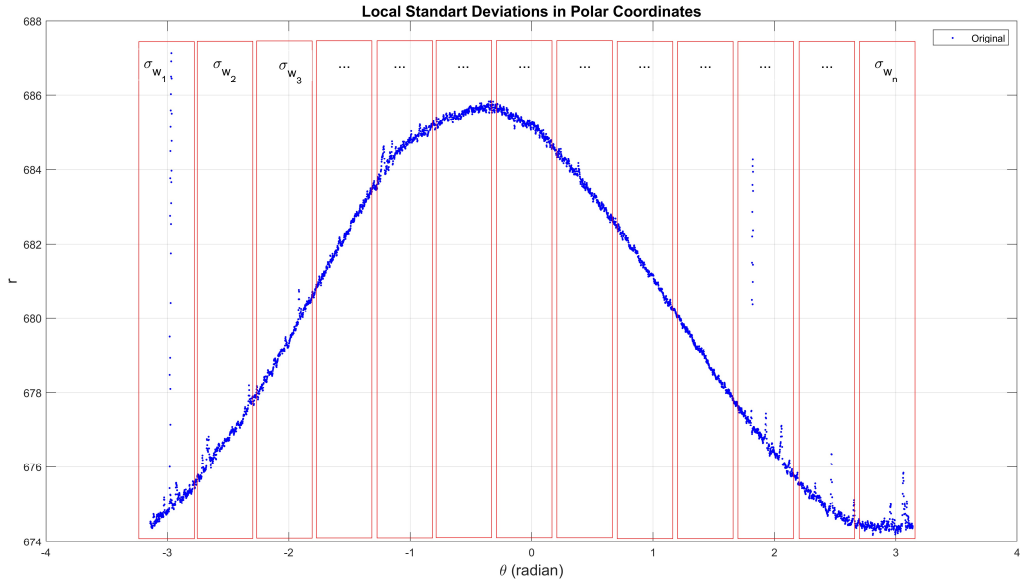

(a)

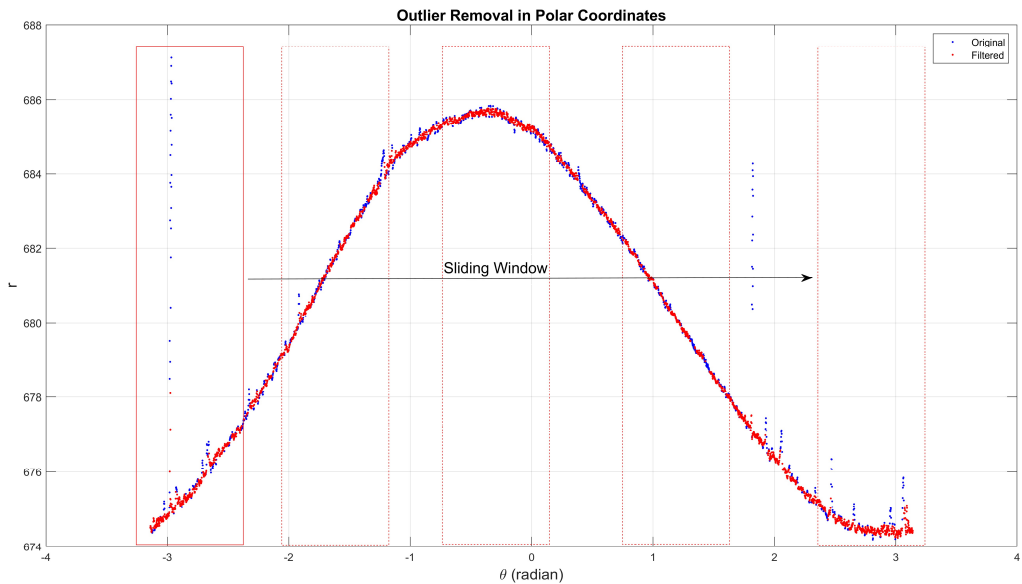

(b)

**Figure 4:** Local filtering of the proposed method in the polar coordinate system. (a) Calculation of the local standard deviation with a fixed stride value, followed by the computation of the global standard deviation. (b) Outlier detection and removal using a sliding window approach.

### 4.4 Circle Fitting

Circle fitting is a critical process in many industrial and scientific applications, where precise measurement of circular features is required. The primary goal of circle fitting is to estimate the parameters of a circle—its center coordinates (a, b) and radius (R)—that best fits a set of edge points. These edge points typically represent the boundary of circular objects, detected using edge detection techniques such as the Canny or Sobel filters. In this study, circle fitting is performed by applying various outlier removal algorithms to the points obtained from the edge detection step. Ten of the most commonly used circle fitting algorithms in the literature—Least Squares Fitting (LSF), Pratt, Taubin, RANSAC, IRLS, HyperLS, M-Estimator, LMedS, TLS, and

EDCircle—are employed. Each of these algorithms offers a unique approach to fitting a circle, from basic algebraic methods to more advanced, robust techniques that handle noisy or inconsistent data points more effectively. The Least Squares Fitting (LSF) method, for example, minimizes the algebraic error between the observed data and the ideal circle. It is simple to implement and commonly used, but can be sensitive to outliers. Pratt's method introduces a geometric constraint to ensure that the fitted circle better represents the true circular geometry, making it more robust than LSF in certain scenarios. Taubin's method further improves upon these by minimizing a first-order geometric approximation of the squared distances between points and the estimated circle, reducing geometric bias. RANSAC (Random Sample Consensus) is a robust algorithm designed for datasets with a significant amount of outliers. It iteratively fits a model to random subsets of the data and selects the best-fit model based on a consensus of inliers. The IRLS (Iteratively Reweighted Least Squares) algorithm, on the other hand, adjusts the weight of each data point iteratively to reduce the influence of outliers. HyperLS introduces constraints to stabilize the solution, enhancing numerical precision. Other algorithms, such as M-Estimator and LMedS, aim to reduce the impact of outliers by minimizing robust loss functions, further increasing the fitting accuracy in the presence of noise. The study investigates whether the proposed method improves the performance of these existing fitting algorithms, particularly in terms of robustness and accuracy in industrial contexts. It also evaluates the comparative performance of fitting methods when combined with different outlier removal algorithms, which are commonly employed to clean noisy edge data before fitting. By combining multiple fitting algorithms with various outlier removal techniques, this study aims to provide a comprehensive performance evaluation framework. The goal is to optimize the circle fitting process in industrial image processing applications, ensuring accurate and reliable results even in challenging environments with noisy data and contaminants.

5. **RESULTS**

The effectiveness of the proposed outlier removal algorithm was evaluated through a comparative study against established methods in the literature. Instead of relying solely on standard least squares fitting, the analysis incorporated ten different circle fitting algorithms to ensure a more comprehensive assessment. A total of 45 industrial workpieces were measured using a Coordinate Measurement Machine (CMM) with an accuracy of ±1 micron, providing ground truth values for validation. The captured images of the workpieces were first processed using an edge detection algorithm, after which various outlier removal methods were applied to eliminate undesired points. The remaining edge points were then used for diameter estimation in pixel units via the selected circle fitting algorithms. The first workpiece was designated as a reference to enable conversion from pixel to millimeter units. For each outlier removal method, a distinct conversion factor was calculated based on the CMM-derived ground truth and the corresponding pixel-based measurement. These conversion factors were then applied to compute the measurements of all workpieces across five outlier removal methods and ten fitting algorithms. Mean Absolute Error (MAE) values were calculated to assess accuracy (Table 1), and standard deviations of the errors were computed to evaluate measurement consistency (Table 2).

Mean Absolute Error

| Outlier Removal | Geometric LS | Pratt | Taubin | RANSAC | IRLS | Hyper LS | M-Estimator | LMedS | TLS | EDCircle |
|---|---|---|---|---|---|---|---|---|---|---|
| Z-Score | 0.0068 | 0.0068 | 0.0068 | 0.0119 | 0.0069 | 0.0068 | 0.0055 | 0.0104 | 0.0892 | 0.0068 |
| MAD | 0.0071 | 0.0071 | 0.0071 | 0.0132 | 0.0071 | 0.0071 | 0.0061 | 0.0093 | 0.0906 | 0.0071 |
| DBSCAN | 0.0065 | 0.0065 | 0.0065 | 0.0116 | 0.0066 | 0.0065 | 0.0055 | 0.0103 | 0.0895 | 0.0065 |
| LOF | 0.0065 | 0.0065 | 0.0065 | 0.0112 | 0.0065 | 0.0065 | 0.0055 | 0.0093 | 0.0895 | 0.0065 |
| Percentile | 0.0067 | 0.0067 | 0.0068 | 0.0147 | 0.0068 | 0.0068 | 0.0055 | 0.0101 | 0.0898 | 0.0068 |
| None | 0.0065 | 0.0065 | 0.0065 | 0.0102 | 0.0066 | 0.0065 | 0.0055 | 0.0100 | 0.0895 | 0.0065 |
| Proposed | 0.0053 | 0.0053 | 0.0053 | 0.0072 | 0.0053 | 0.0053 | 0.0054 | 0.0088 | 0.0910 | 0.0053 |

**Table 1**. Mean Absolute Error (MAE) comparison of various circle fitting algorithms under different outlier removal methods.

Standart Deviation of Absolute Error

| Outlier Removal | Geometric LS | Pratt | Taubin | RANSAC | IRLS | Hyper LS | M-Estimator | LMedS | TLS | EDCircle |
|---|---|---|---|---|---|---|---|---|---|---|
| Z-Score | 0.0036 | 0.0036 | 0.0036 | 0.0087 | 0.0035 | 0.0036 | 0.0035 | 0.0103 | 0.0475 | 0.0036 |
| MAD | 0.0044 | 0.0044 | 0.0044 | 0.0092 | 0.0043 | 0.0044 | 0.0046 | 0.0080 | 0.0470 | 0.0044 |
| DBSCAN | 0.0033 | 0.0033 | 0.0033 | 0.0093 | 0.0032 | 0.0033 | 0.0035 | 0.0087 | 0.0476 | 0.0033 |
| LOF | 0.0033 | 0.0033 | 0.0033 | 0.0080 | 0.0032 | 0.0033 | 0.0035 | 0.0089 | 0.0476 | 0.0033 |
| Percentile | 0.0036 | 0.0036 | 0.0036 | 0.0090 | 0.0037 | 0.0036 | 0.0035 | 0.0096 | 0.0476 | 0.0036 |
| None | 0.0033 | 0.0033 | 0.0033 | 0.0078 | 0.0032 | 0.0033 | 0.0035 | 0.0097 | 0.0476 | 0.0033 |
| Proposed | 0.0035 | 0.0035 | 0.0035 | 0.0060 | 0.0035 | 0.0035 | 0.0036 | 0.0085 | 0.0477 | 0.0035 |

**Table 2**. Standart Deviation of Absolute Error comparison of various circle fitting algorithms under different outlier removal methods.

The results, presented in Table 1, show a detailed comparison of the performance of various circle fitting algorithms combined with different outlier removal techniques. The Mean Absolute Error (MAE) values reveal that the Proposed Method significantly outperforms all other outlier removal techniques across all circle fitting algorithms. This method achieves the lowest MAE values, particularly for algorithms like Geometric LS, Pratt, Taubin, IRLS, and Hyper LS, with consistent MAE values of 0.0053. These results clearly demonstrate the effectiveness of the proposed method in improving the accuracy of circle fitting, as it consistently reduces error compared to traditional methods. In comparison, methods like Z-Score, MAD, and DBSCAN perform relatively well, with MAE values around 0.0065 to 0.0068. However, they still fall short when compared to the Proposed Method, particularly in cases involving more complex fitting algorithms such as Pratt and Taubin, where the error level increases, indicating the superiority of the proposed technique in handling outliers effectively. Furthermore, when evaluating the Standard Deviation of Absolute Error (SDAE), which reflects the consistency of the measurements, the Proposed Method continues to lead, maintaining the lowest variability in error, with an SDAE of 0.0035. This showcases the stability of the measurements when the proposed outlier removal approach is applied. Z-Score and MAD also perform well, with an SDAE of 0.0036 to 0.0044, but they exhibit higher variability compared to the Proposed Method. Techniques such as DBSCAN and LOF also provide solid results, with an SDAE range of 0.0033 to 0.0035, but still lag behind the proposed method in terms of error consistency. The Percentile method, while effective in removing outliers, results in higher error variability, with an SDAE ranging from 0.0036 to 0.0037, indicating that its performance is less stable, especially in more complex scenarios. The None category, where no outlier removal is applied, yields the highest error variability with an SDAE range from 0.0033 to 0.0097, emphasizing the importance of outlier removal for achieving reliable and stable measurements. These findings reinforce the critical role of outlier removal techniques in circle fitting tasks. The Proposed Method stands out as the most effective approach, providing both higher accuracy and greater consistency in measurements, particularly when compared to traditional outlier detection methods like Z-Score, MAD, and DBSCAN. This improved performance is essential for industrial applications that require high precision and stability, such as in the manufacturing and quality control processes. By reducing both the MAE and SDAE, the proposed method offers a reliable solution for accurate circle fitting, making it a valuable tool in industrial image processing systems.

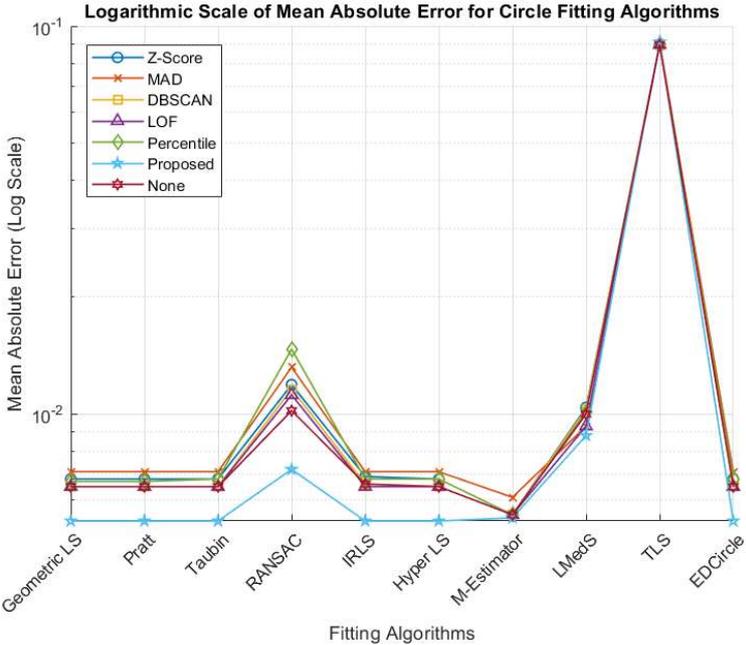

**Figure 5.** Logarithmic plot of the Mean Absolute Error (MAE) for different circle fitting algorithms combined with various outlier removal methods.

Figure 5 illustrates the logarithmic visualization of Mean Absolute Error (MAE) across various circle fitting algorithms when combined with different outlier removal techniques. Unlike Table 1, which presents exact numerical comparisons, this graph highlights relative performance trends and emphasizes the stability and superiority of the proposed method across all fitting algorithms. The log scale effectively reveals subtle differences among the lower MAE values, making it easier to observe performance distinctions between techniques such as Z-Score, MAD, DBSCAN, and the proposed method. Notably, the Proposed Method consistently maintains the lowest MAE values with minimal variation, especially in algorithms like RANSAC and LMedS, where other methods exhibit a sharp increase in error. The visual representation further underscores the robustness of the proposed approach, particularly in scenarios involving high variance or model sensitivity, and supports its applicability in precision-critical industrial environments.

## 6. CONCLUSION

In this study, a novel outlier removal algorithm based on polar coordinates is proposed, which can be applied prior to circle fitting algorithms. The suggested method takes the edge points obtained from the edge detection step as input and initially transforms these points into the polar coordinate system. Following this transformation, a local window-based standard deviation approach is employed to calculate the global standard deviation. This value is then used as a threshold to detect outliers by evaluating each point against the local mean radius and global standard deviation within the same local window. Outlier points, which deviate significantly from the expected circular structure, are identified and filtered out to enhance the accuracy of the subsequent circle fitting process. For the experiments, a dataset consisting of 45 industrial-grade washer-shaped workpieces was used. The images of the workpieces were captured using an industrial camera system designed for high-precision measurements. The performance of the proposed Polar Coordinate-Based Outlier Detection (PCOD) algorithm was compared with several traditional outlier detection methods. The results indicate that the proposed method significantly outperforms the other methods in terms of accuracy. Specifically, the PCOD algorithm demonstrated superior performance in reducing the mean absolute error (MAE), improving the fitting results of nine out of the ten circle fitting algorithms evaluated in this study. This demonstrates the effectiveness of the proposed method in enhancing the precision of circle fitting, especially in industrial applications where high accuracy and reliability are crucial. In particular, the PCOD method outperformed the other outlier removal techniques by achieving more stable and consistent results, which is essential for industrial environments where components are often subject to noise and contamination. The ability of the proposed algorithm to effectively filter out outliers prior to circle fitting contributes to a more accurate estimation of geometric parameters such as the center and radius of circular objects. This improvement in performance highlights the potential of the PCOD algorithm to significantly enhance the reliability and precision of industrial measurement systems, particularly in high-precision applications such as manufacturing and quality control. Furthermore, the proposed algorithm provides a robust solution for dealing with noisy and inconsistent data, ensuring more reliable measurements in challenging industrial environments. In summary, the PCOD algorithm not only provides a more effective approach for outlier detection but also offers a significant advancement in the accuracy and stability of circle fitting applications. The promising results from this study underscore the potential of the proposed method for improving the quality of dimensional measurements in industrial image processing systems, marking a valuable contribution to the existing literature on outlier removal and circle fitting methodologies.

**Acknowledgments**
The author gratefully acknowledges Dogu Pres Automotive for their invaluable support in facilitating this research.

**References**

[1] Chernov, Nikolai, and Claire Lesort. "Least squares fitting of circles." *Journal of Mathematical Imaging and Vision* 23 (2005): 239-252. https://doi.org/10.1007/s10851-005-0482-8

[2] Tamura, Hajime, et al. "Circle fitting based position measurement system using laser range finder in construction fields." *2010 IEEE/RSJ International Conference on Intelligent Robots and Systems*. IEEE, 2010. https://doi.org/10.1109/IROS.2010.5649211

[3] Lin, Shan, et al. "A least squares algorithm for fitting data points to a circular arc cam." *Measurement* 102 (2017): 170-178. https://doi.org/10.1016/j.measurement.2017.01.059

[4] Lee, Byung-Ryong, and Huu-Cuong Nguyen. "Development of laser-vision system for three-dimensional circle detection and radius measurement." *Optik* 126.24 (2015): 5412-5419. https://doi.org/10.1016/j.ijleo.2015.09.131

[5] Pratt, Vaughan. "Direct least-squares fitting of algebraic surfaces." *ACM SIGGRAPH computer graphics* 21.4 (1987): 145-152. https://doi.org/10.1145/37402.37420

[6] Taubin, Gabriel. "An improved algorithm for algebraic curve and surface fitting." *1993 (4th) International Conference on Computer Vision*. IEEE, 1993. https://doi.org/10.1109/ICCV.1993.378149

[7] FISCHLER AND, M. A. "Random sample consensus: a paradigm for model fitting with applications to image analysis and automated cartography." *Commun. ACM* 24.6 (1981): 381-395. https://doi.org/10.1145/358669.358692

[8] Guo, Jianfeng, and Jiameng Yang. "An iterative procedure for robust circle fitting." *Communications in Statistics-Simulation and Computation* 48.6 (2019): 1872-1879. https://doi.org/10.1080/03610918.2018.1425443

[9] Kanatani, Kenichi, and Prasanna Rangarajan. "Hyper least squares fitting of circles and ellipses." *Computational Statistics & Data Analysis* 55.6 (2011): 2197-2208. https://doi.org/10.1016/j.csda.2010.12.012

[10] Guo, Jianfeng, and Jiameng Yang. "An iterative procedure for robust circle fitting." *Communications in Statistics-Simulation and Computation* 48.6 (2019): 1872-1879. https://doi.org/10.1080/03610918.2018.1425443

[11] Umbach, Dale, and Kerry N. Jones. "A few methods for fitting circles to data." *IEEE Transactions on instrumentation and measurement* 52.6 (2003): 1881-1885. https://doi.org/10.1109/TIM.2003.820472

[12] Ester, Martin, et al. "A density-based algorithm for discovering clusters in large spatial databases with noise." *kdd*. Vol. 96. No. 34. 1996. https://doi.org/10.1145/3001460.3001507

[13] Akinlar, Cuneyt, and Cihan Topal. "EDCircles: A real-time circle detector with a false detection control." *Pattern Recognition* 46.3 (2013): 725-740. https://doi.org/10.1016/j.patcog.2012.09.020

[14] Huber, Peter J. "Robust statistics." *International encyclopedia of statistical science*. Springer, Berlin, Heidelberg, 2011. 1248-1251. https://doi.org/10.1007/978-3-642-04898-2_594

[15] Iglewicz, Boris, and David C. Hoaglin. *Volume 16: how to detect and handle outliers*. Quality Press, 1993.

[16] Ester, Martin, et al. "A density-based algorithm for discovering clusters in large spatial databases with noise." *kdd*. Vol. 96. No. 34. 1996.

[17] Breunig, Markus M., et al. "LOF: identifying density-based local outliers." *Proceedings of the 2000 ACM SIGMOD international conference on Management of data*. 2000. https://doi.org/10.1145/342009.335388

[18] Rousseeuw, Peter J., and Annick M. Leroy. *Robust regression and outlier detection*. John wiley & sons, 2003.

[19] Poyraz, Ahmet Gökhan, et al. "Sub-Pixel counting based diameter measurement algorithm for industrial Machine vision." *Measurement* 225 (2024): 114063. https://doi.org/10.1016/j.measurement.2023.114063

[20] von Gioi, Rafael Grompone, and Gregory Randall. "A sub-pixel edge detector: an implementation of the canny/devernay algorithm." *Image Processing On Line* 7 (2017): 347-372. https://doi.org/10.5201/ipol.2017.216

[21] Trujillo-Pino, Agustín, et al. "Accurate subpixel edge location based on partial area effect." *Image and vision computing* 31.1 (2013): 72-90. https://doi.org/10.1016/j.imavis.2012.10.005

[22] Peng, Gaoliang, Zhujun Zhang, and Weiquan Li. "Computer vision algorithm for measurement and inspection of O-rings." *Measurement* 94 (2016): 828-836. https://doi.org/10.1016/j.measurement.2016.09.012

[23] Da, Feipeng, and Hu Zhang. "Sub-pixel edge detection based on an improved moment." *Image and Vision Computing* 28.12 (2010): 1645-1658. https://doi.org/10.1016/j.imavis.2010.05.003

[24] Chu, Menglin, et al. "Sub-pixel dimensional measurement algorithm based on intensity integration threshold." *OSA Continuum* 3.10 (2020): 2912-2924. https://doi.org/10.1364/OSAC.402101

[25] Xie, Xin, et al. "An improved industrial sub-pixel edge detection algorithm based on coarse and precise location." *Journal of Ambient Intelligence and Humanized Computing* 11 (2020): 2061-2070. https://doi.org/10.1007/s12652-019-01232-2

[26] Sun, Qiucheng, et al. "A robust edge detection method with sub-pixel accuracy." *Optik* 125.14 (2014): 3449-3453. https://doi.org/10.1016/j.ijleo.2014.02.001

[27] Ying-Dong, Qu, et al. "A fast subpixel edge detection method using Sobel–Zernike moments operator." *Image and Vision Computing* 23.1 (2005): 11-17. https://doi.org/10.1016/j.imavis.2004.07.003

[28] Ghosal, Sugata, and Rajiv Mehrotra. "Orthogonal moment operators for subpixel edge detection." *Pattern recognition* 26.2 (1993): 295-306. https://doi.org/10.1016/0031-3203(93)90038-X